\documentclass[10pt,twocolumn,letterpaper]{article}

\usepackage[review]{cvpr}      % To produce the REVIEW version

\usepackage[dvipsnames]{xcolor}

\definecolor{cvprblue}{rgb}{0.21,0.49,0.74}
\usepackage[pagebackref,breaklinks,colorlinks,citecolor=cvprblue]{hyperref}
\usepackage{floatrow}
\title{FusionFrames: Efficient Architectural Aspects for\\ Text-to-Video Generation Pipeline}

\author{Vladimir Arkhipkin$^1$
\and
Zein Shaheen$^1$
\and
Viacheslav Vasilev$^{1,2}$
\and
Elizaveta Dakhova$^3$
\and
Andrey Kuznetsov$^{1,3}$
\and
Denis Dimitrov$^{1,3}$
\and
\;\;$^1$Sber AI \;\;\;\;\; $^2$Moscow Institute of Physics and Technology \;\;\;\;\; $^3$Artificial Intelligence Research Institute\\
{\tt\small \{kuznetsov, dimitrov\}@airi.net}
}

\begin{document}
\maketitle
\begin{abstract}
Multimedia generation approaches occupy a prominent place in artificial intelligence research. Text-to-image models achieved high-quality results over the last few years. However, video synthesis methods recently started to develop. This paper presents a new two-stage latent diffusion text-to-video generation architecture based on the text-to-image diffusion model. The first stage concerns keyframes synthesis to figure the storyline of a video, while the second one is devoted to interpolation frames generation to make movements of the scene and objects smooth. We compare several temporal conditioning approaches for keyframes generation. The results show the advantage of using separate temporal blocks over temporal layers in terms of metrics reflecting video generation quality aspects and human preference. The design of our interpolation model significantly reduces computational costs compared to other masked frame interpolation approaches. Furthermore, we evaluate different configurations of MoVQ-based video decoding scheme to improve consistency and achieve higher PSNR, SSIM, MSE, and LPIPS scores. Finally, we compare our pipeline with existing solutions and achieve top-2 scores overall and top-1 among open-source solutions: CLIPSIM = 0.2976 and FVD = 433.054. Code is available here: \url{https://github.com/ai-forever/KandinskyVideo}.
\end{abstract}    
\section{Introduction}

Text-to-image (T2I) generation approaches have achieved stunning results in recent years \cite{Nichol2022glide, ramesh2022hierarchical, rombach2022high, saharia2022photorealistic}. The task of video generation is a natural and logical continuation of the development of this direction. Diffusion probabilistic models \cite{Dickstein2015, ho2020denoising, Song2021SBD} played an essential role in image generation quality improvement. Text-to-video (T2V) generative diffusion models are also becoming extremely popular, but the problems inherent in this task still pose a severe challenge.

Such problems include training and inference computational costs and require large, high-quality, open-source text+video datasets. The available data is insufficient to fully understand all the generation possibilities when training from scratch. In addition, such datasets impose restrictions on models related to the specificity of video domain. Furthermore, to achieve high realism and aestheticism, video generation requires not only the visual quality of a single frame but also the frame coherence in terms of semantic content and appearance, smooth transitions of objects on adjacent frames, and correct physics of movements. The primary key for the mentioned aspects is the temporal information which is essential to the video modality and represents space-time correlations. Hence, the generation quality will largely depend on the data processing along the time dimension of video sequences.

As a rule, temporal information is taken into account in diffusion models by including temporal convolutional layers or temporal attention layers in the architecture \cite{ho2022video, wu2022tuneavideo, singer2022make, ho2022imagen, esser2023gen1, blattmann2023align, zhou2022magicvideo, zhang2023controlvideo, li2023videogen}. This allows initializing the weights of the remaining spatial layers with the weights of the pretrained T2I model and training only the temporal layers.
In this way, we can reduce the necessity for large-scale text-video pairs datasets because we can transfer comprehensive knowledge of T2I models to the video domain. Also, using latent diffusion models \cite{rombach2022high} further reduces the computational costs.

In this paper, we present our T2V generation architecture based on latent diffusion and examine various architectural aspects to enhance the overall quality, consistency and smoothness of generated videos. The proposed pipeline is divided into two stages: the keyframe generation stage, designed to control the video main storyline, and the interpolation stage regarding movement smoothness improvement by generating additional frames. This separation allows us to maintain alignment with the text description throughout the entire video in terms of both content and dynamics. At the keyframe generation stage, we compare temporal conditioning approaches, namely traditional mixed spatial-temporal blocks and three types of \textit{separate temporal blocks}. We find that using separate temporal blocks significantly improves the video quality, which is supported by the quality metrics and human evaluation study. We propose this solution as a general approach to include temporal components in T2I models to use them for video generation. For our interpolation model, we design the architecture to reduce inference running time by predicting a group of interpolated frames together instead of individual frames, and we find that it also improves the quality of interpolated frames. Regarding the latent representation of frames, we comprehensively analyze various options for constructing the MoVQGAN-based \cite{zheng2022movq} video decoder, assessing them in terms of quality metrics and the number of additional parameters. This analysis is aimed at enhancing the consistency of adjacent frames throughout the video.

Thus, our contribution contains the following aspects:
\begin{itemize}
    \item We present FusionFrames -- the end-to-end T2V latent diffusion pipeline, which is based on the pretrained frozen T2I model Kandinsky 3.0 \cite{arkhipkin2023kandinsky}. The pipeline is divided into two parts -- key frames generation and interpolation frames synthesis.
    \item As a part of the keyframes generation, we propose to use separate temporal blocks for processing temporal information. We compare three types of blocks with mixed spatial-temporal layers and demonstrate the qualitative and quantitative advantage of our solution in terms of visual appearance and temporal consistency using a set of metrics (FVD, IS, CLIPSIM) and human evaluation study on several video datasets in different domains. The conducted experiments show top-2 performance in terms of CLIPSIM and FVD scores regarding other published results — 0.2976 and 433.054 respectively. 
    \item We present an efficient interpolation architecture that runs more than three times faster compared to other popular masked frame interpolation architectures and generates interpolated frames with higher fidelity.
    \item We investigate various architectural options to build MoVQ-GAN video decoder and evaluate their performance in terms of quality metrics and the impact on the size of decoder.
\end{itemize}
    \section{Related Work}
    \subsection{Text-to-Video Generation}

    Prior works on video generation utilize VAEs \cite{mittal2017sync, babaeizadeh2017stochastic, babaeizadeh2021fitvid, yan2021videogpt, walker2021predicting}, GANs \cite{Vondrick2016Videos, pan2017create, li2018video, lee2018stochastic, clark2019adversarial}, normalizing flows \cite{kumar2019videoflow} and autoregressive transformers \cite{wu2021godiva, wu2021nuwa, ge2022long, hong2022cogvideo, Villegas2022Phenaki}. GODIVA \cite{wu2021godiva} adopts a 2D VQVAE and sparse attention for T2V generation. CogVideo \cite{hong2022cogvideo} is built on top of a frozen CogView2 \cite{ding2022cogview2} T2I transformer by adding additional temporal attention layers.

    Recent research extend T2I diffusion-based architecture for T2V generation \cite{singer2022make, ho2022imagen, blattmann2023align, he2022latent, zhou2022magicvideo}. This approach can benefit from pretrained image diffusion models and transfer that knowledge to video generation tasks. Specifically, it introduces temporal convolution and temporal attention layers interleaved with existing spatial layers. This adaptation aims to capture temporal dependencies between video frames while achieving computational efficiency by avoiding infeasible 3D convolutions and 3D attention mechanisms. These temporal layers can be trained independently \cite{singer2022make, blattmann2023align} or jointly \cite{ho2022imagen, esser2023gen1} with the 2D spatial layers.
    This technique of mixed spatial-temporal blocks for both convolutions and attention layers has become widespread in most T2V models \cite{ho2022video, singer2022make, ho2022imagen, esser2023gen1, wu2022tuneavideo, blattmann2023align, zhou2022magicvideo, li2023videogen}. Alternative approaches to operating with the time dimension include the use of an image diffusion model in conjunction with a temporal autoregressive Recurrent Neural Network (RNN) to predict individual video frames \cite{yang2022diffusion}, projection of 3D data into a latent 2D space \cite{yu2023video} and the use of diffusion to generate latent flow sequences \cite{ni2023conditional}. In this paper, we propose separate temporal blocks as a new approach for conditioning temporal information.

    Finally, it is worth noting that certain studies \cite{singer2022make, ho2022imagen} operate entirely in pixel space, whereas others \cite{he2022latent, zhou2022magicvideo, wu2022tuneavideo, esser2023gen1, blattmann2023align, zhang2023controlvideo, li2023videogen} utilize the more efficient latent space. We follow the second approach in this paper.
        
    \subsection{Video Frame Interpolation}

\begin{figure*}[ht]
    \center{\includegraphics[scale=0.12]{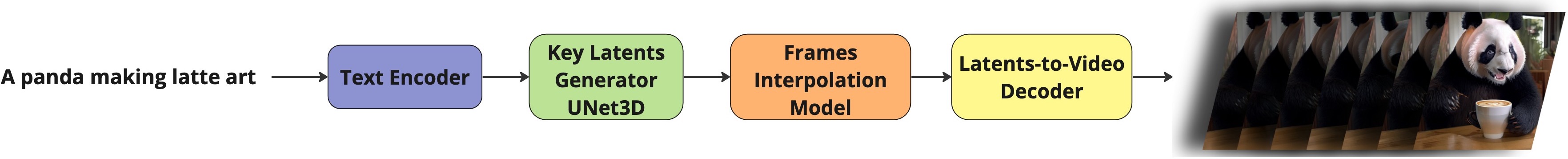}}
    \caption{{\textbf{Overall scheme of our pipeline.} The encoded text prompt enters the U-Net keyframe generation model with temporal layers or blocks, and then the sampled latent keyframes are  sent to the latent interpolation model in such a way as to predict three interpolation frames between two keyframes. A temporal MoVQ-GAN decoder is used to get the final video result.}}
    \label{fig:pipeline}
\end{figure*} 
        
MCVD~\cite{voleti2022mcvd} uses a diffusion-based model for interpolation. It leverages four consecutive keyframes (two from each side) to predict three frames in between.
    In the field of T2V research, diffusion-based architectures are commonly adopted for interpolation. 
    Some studies make use of training methods involving \emph{Masked Frame Interpolation}~\cite{ho2022imagen,singer2022make,blattmann2023align,zhang2023show}. 
    In LVDM~\cite{he2022latent}, they use \emph{Masked Latent Clip Interpolation} instead.
    VideoGen~\cite{li2023videogen} employs a \emph{flow-based approach} for interpolation.
    Magicvideo~\cite{zhou2022magicvideo} combines the latents of two consecutive keyframes with randomly sampled noise. 
    It is worth noting that several training techniques have been proposed, including \emph{Conditional Frame (or Latent) Perturbation}~\cite{ho2022imagen,he2022latent,zhang2023show} to mitigate the error introduced by the previous generation step and \emph{Context Guidance (or Unconditional Guidance)}~\cite{blattmann2023align,he2022latent} to enhance the diversity and fidelity of sampled video.
    
    In our interpolation network, we adjust the input and output layers in U-Net to generate three identical frames before training. Then, we train U-Net to predict three interpolated frames (between each two consecutive keyframes). This adaptation significantly reduces the computational cost compared to \emph{Masked Frame Interpolation} methods, enabling pre-trained T2I weights for initialization.
    
\subsection{Video Decoder}
 
    In their works, \cite{blattmann2023align, li2023videogen} builds a video decoder with temporal attention and convolution layers. MagicVideo~\cite{zhou2022magicvideo}, on the other hand, incorporates two temporal-directed attention layers in the decoder to build a VideoVAE. To the best of our knowledge, previous studies have not compared their strategies for constructing a video decoder. In this research, we present multiple options for designing a video decoder and conduct an extensive comparative analysis, evaluating their performance regarding quality metrics and the implications for additional parameters.

\section{Diffusion Probabilistic Models}
\label{sec:dpm}

  Denoising Diffusion Probabilistic Models (DDPM) \cite{ho2020denoising} is a family of generative models designed to learn a target data distribution $p_{data}(x)$. It consists of a forward diffusion  process and a backward denoising process. In the forward process, random noise is gradually added into the data $x$ through a $T$-step Markov chain \cite{kong2021fast}. The noisy latent variable at step $t$ can be expressed as:
    
  \begin{equation}
    z_t = \sqrt{\hat{\alpha_t}}x+\sqrt{1-\hat{\alpha_t}}\epsilon
  \end{equation}

  with 
    $\hat{\alpha}_t = \prod_{k=1}^{t}\alpha_k, 
    0 \le {\alpha}_k < 1, 
    \epsilon \sim N(0,1)$. For a sufficiently large $T$, e.g., $T = 1000$, $\sqrt{\hat{\alpha}_T}\approx 0$, and \\$1 - \sqrt{\hat{\alpha}_T}\approx 1$. Consequently, $z_T$  ends up with pure noise. The generation of $x$ can then be seen as an iterative denoising process. This denoising process corresponds to learning the inverted process of a fixed Markov Chain of length T.

    \begin{equation}
    L_t(x) = \mathbb{E}_{\epsilon \sim N(0,1)}[\|\epsilon - z_{\theta}(z_t, t)\|_2^2]
    \end{equation}

    Here, $z_{\theta}$ represents a denoising neural network parameterized by $\theta$, and $L_t$ is the loss function.
\section{Methods}

\begin{figure*}[ht]
    \center{\includegraphics[scale=0.2]{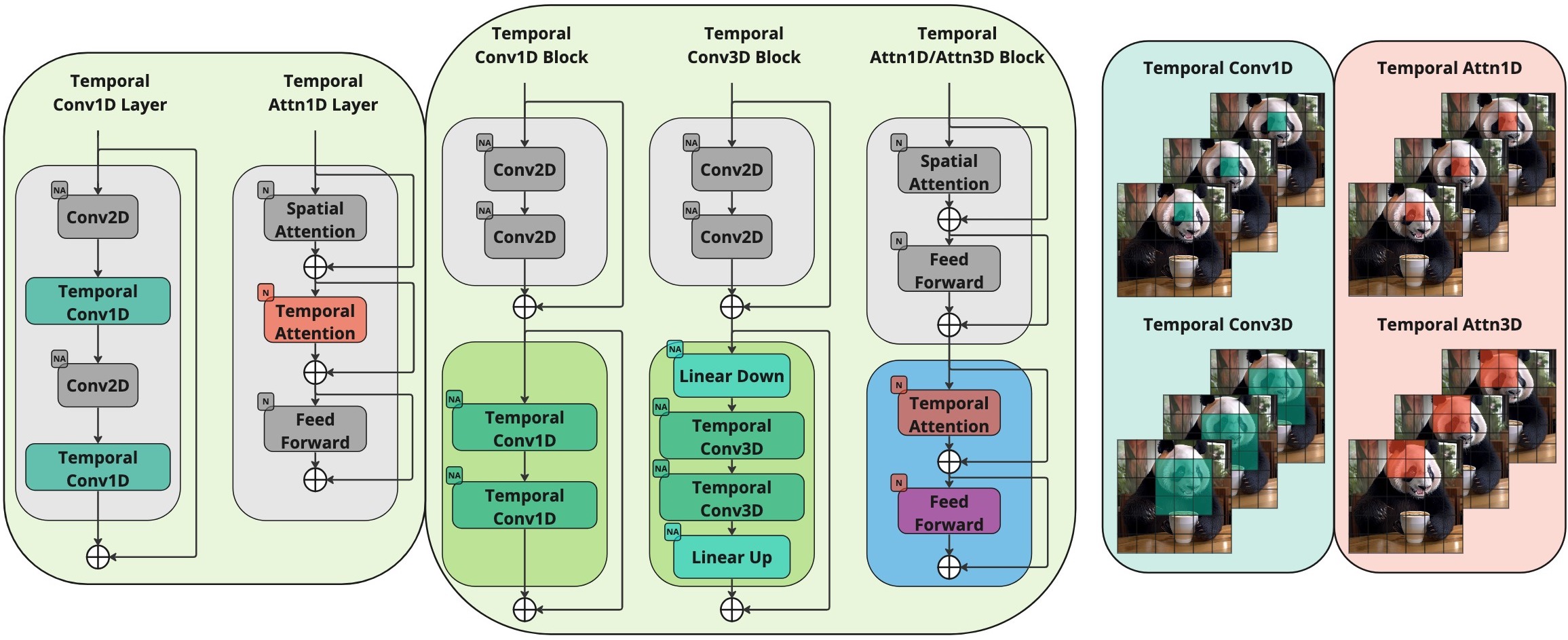}} \\
    \caption{\textbf{Comparative schemes for temporal conditioning.} We examined two approaches of temporal components use in pretrained architecture of T2I U-Net from Kandinsky 3.0 \cite{arkhipkin2023kandinsky} -- the traditional approach of mixing spatial and temporal layers in one block \textbf{(left)} and our approach of allocating a separate temporal block \textbf{(middle)}. All layers indicated in gray are not trained in T2V architectures and are initialized with the weights of the T2I Kandinsky 3.0 model. NA and N in the left corner of all layers correspond to the presence of prenormalization layers with and without activation, respectively. For different types of blocks we implemented different types of temporal attention and temporal convolution layers, \textbf{(left)}. We also implement different types of temporal conditioning. One of them is simple conditioning when pixel see only itself value in different moments of type (1D layers). In 3D layers pixels can see the values of its neighbors also in different moments of time \textbf{(right)}.}
    \label{fig:conv_and_attn_scheme}
\end{figure*}

\paragraph{Overall pipeline.} The scheme of our T2V pipeline is shown in Figure \ref{fig:pipeline}. It includes a text encoder, a keyframe latent generation model, a frame interpolation model, and a latent decoder. Below, we describe these key components in details.

\subsection{Keyframes Generation with Temporal Conditioning}

The keyframes generation is based on a pretrained latent diffusion T2I model Kandinsky 3.0 \cite{arkhipkin2023kandinsky}. We use the weights of this model to initialize the spatial layers of the keyframe generation model, which is distinguished by the presence of temporal components. In all experiments, we freeze the weights of T2I U-Net and train only temporal components. We consider two fundamental ways of introducing temporal components into the general architecture -- temporal layers of convolutions with attention integrated into spatial blocks and our separate temporal blocks. Figure \ref{fig:conv_and_attn_scheme} exhaustively explains our concept.

We also investigate different types of temporal conditioning in convolution and attention layers. In the case of convolution, we use its 3D version with kernels $3 \times 1 \times 1$ and $3 \times 3 \times 3$ corresponding to \textit{Temporal Conv1D} and \textit{Temporal Conv3D}, respectively. A similar mechanism was implemented for temporal attention layers to build interaction not only between latent positions between each other across time dimension (\textit{Temporal Attn1D}) but also between a window of positions across the video (\textit{Temporal Attn3D}) (Figure~\ref{fig:conv_and_attn_scheme}~-~right). We used an attention window of size $2 \times 2 \times T$ in Temporal Attn3D, where $T$ is the total number of generated frames. The algorithm of dividing the whole frame into groups of four pixels is the same as Block Attention in MaxViT \cite{tu2022maxvit}.

In our case, separate temporal blocks are organized in three ways:
\begin{itemize}
    \item \emph{Conv1dAttn1dBlocks} -- Incorporation of a Temporal Conv1D Block following each spatial convolution block and a Temporal Attn1D Block following each self-attention block in the T2I model.
    \item \emph{Conv3dAttn1dBlocks} -- Incorporation of a Temporal Conv3D Block following each spatial convolution block and a Temporal Attn1D Block following each self-attention block. Additionally, two extra Linear layers are introduced within the Temporal Conv3D Block to downsample the hidden dimension of the input tensor and subsequently upsample it to the original size. These projections are employed to maintain a comparable number of parameters as in the \emph{Conv1dAttn1dBlocks} configuration.
    \item \emph{Conv1dAttn3dBlocks} -- Incorporation of a Temporal Conv1D Block following each spatial convolution block and a Temporal Attn3D Block following each self-attention block. In this configuration, no extra parameters are introduced compared to \emph{Conv1dAttn1dBlocks}, as the alteration is limited to the sequence length in attention calculation. Moreover, there is minimal computational overhead due to the relatively low number of keyframes (16 for all our models)..
\end{itemize}

In the case of the mixed spatial-temporal block (\emph{Conv1dAttn1dLayers}), we exclusively employ the most standard form of integration, incorporating Temporal Conv1D and Temporal Attn1D layers into the inner section of the spatial blocks (Figure \ref{fig:conv_and_attn_scheme} - left). We made this choise because the integration style represented by \emph{Conv1dAttn1dLayers} consistently yields worse results across all metrics when compared to \emph{Conv1dAttn1dBlocks}.

\subsection{Video Frame Interpolation}
  \label{sec:interpolation}
\begin{figure}[t]
    \center{\includegraphics[scale=0.23]{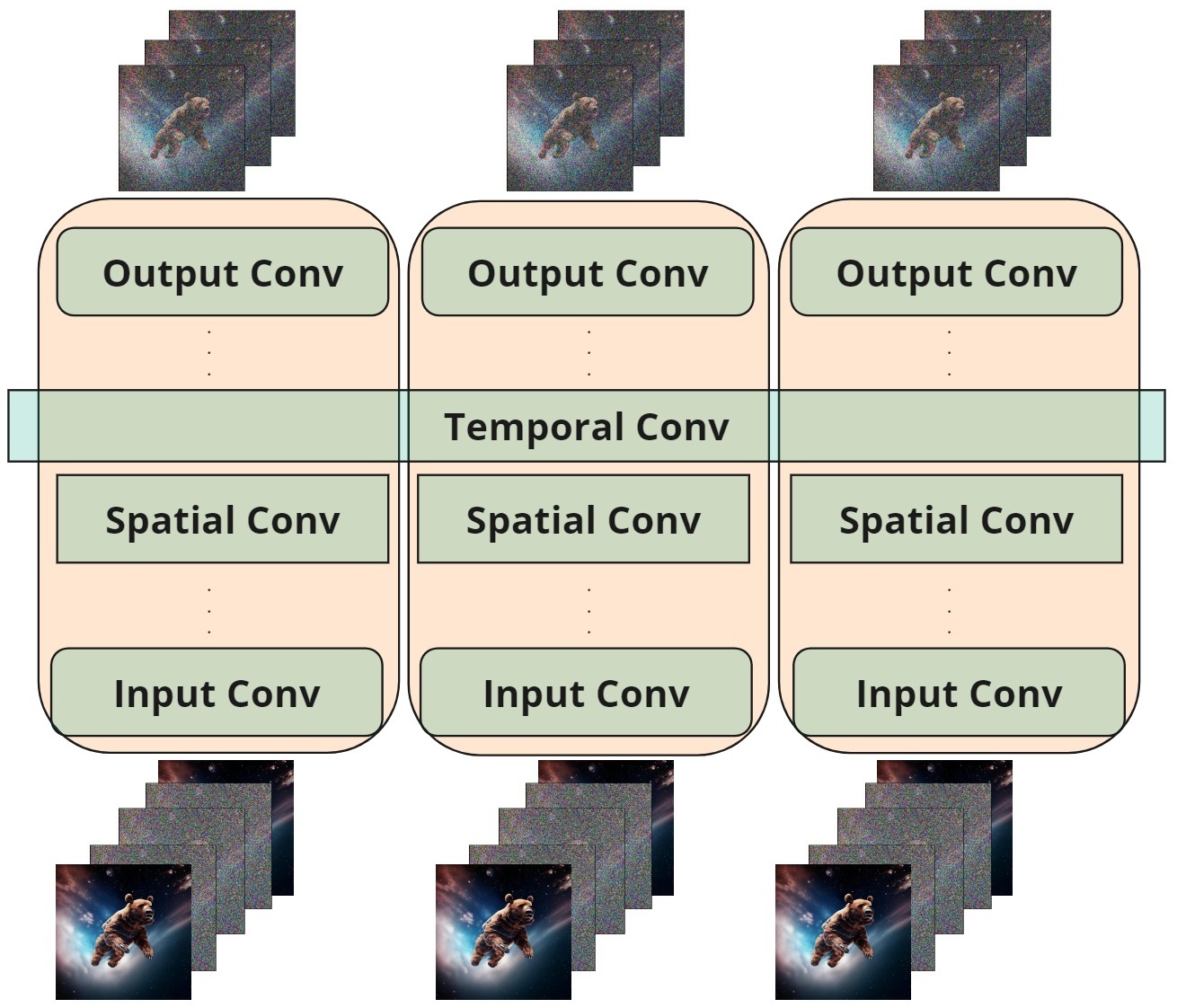}}
    \caption{\textbf{FusionFrames Interpolation Architecture.} A summary of the primary changes made to the T2I architecture includes: (i) The input convolution now accepts three noisy latent inputs (interpolated frames) and two conditioning latent inputs (keyframes). (ii) The output convolution predicts three denoised latents. (iii) Temporal convolutions have been introduced after each spatial convolution.}
    \label{fig:interpolation_arch}
\end{figure}

  We apply interpolation in the latent space to predict a group of three frames between each pair of consecutive keyframes. 
  This necessitates the adaptation of the T2I architecture (Figure~\ref{fig:interpolation_arch}).
  First, we inflate the input convolution layer (by zero-padding weights) to process: (i) A group of three noisy latents representing interpolated frames (${z_t=[z_t^1,z_t^2,z_t^3];z_t^i\in R^{4\times32\times32}}$),
  and (ii) two conditioning latents representing keyframes (${c=[c^1,c^2];c^i\in R^{4\times32\times32}}$).
  The input to U-Net consists of the channel-wise concatenation of $z_t$ and $c$ (${[z_t,c]\in R^{20\times32\times32}}$). 
  Similarly, we inflate the output convolution layer (by replicate padding weights) to predict a group of three denoised latents (${z_{t-1}=[z_{t-1}^1,z_{t-1}^2,z_{t-1}^3];z_{t-1}\in R^{12\times32\times32}}$). 
  Secondly, we insert temporal convolution layers $\phi$ in the original T2I model $\theta$. 
  This enables upsampling a video with arbitrary length $T\rightarrow4T-3$ frames.
  The activation derived from a temporal layer $out_{temporal}$ is combined with the output of the corresponding spatial layer $out_{spatial}$ using a trainable parameter $\alpha$ as follows:
    \begin{equation}
        \label{eq.merge}
        out = out_{spatial} + \alpha \cdot out_{temporal}
    \end{equation}
  A detailed description of these adjustments is provided in Appendix~\ref{appendix_interpolation_modifications}. 
  Finally, we remove text conditioning, and instead, we incorporate skip-frame $s$ (section~\ref{subsec_experimental_setup}) and perturbation level $tp$ conditioning (Check below for details). For interpolation, we use v-prediction parameterization ($v_t \equiv \alpha_t \epsilon-\sigma_t x$) as described in \cite{salimans2022progressive, ho2022imagen}.

  % not clear 
  Building on the conditional frame perturbation technique  \cite{ho2022imagen, blattmann2023align, he2022latent, zhang2023show}, we randomly sample a perturbation level (number of forward diffusion steps) $pt \in \{0, 1, 2, \ldots, 250\}$ and use it to perturb the conditioning latents $c$ using forward diffusion process (Section~\ref{sec:dpm}). This perturbation level is also employed as a condition for the U-Net (as described earlier). 

  With some probability $uncond\_prob$, we also replace the conditioning latents $c$ with zeros for unconditional generation training. The final training objective looks like:
  \begin{equation}
    \begin{split}
      L_t(x;c,&tp,s,m) =\\ &\mathbb{E}_{\epsilon \sim N(0,1)} [\|v_t - z_{\theta, \phi}([z_t,c], t, tp, s, m)\|_2^2]
    \end{split}
  \end{equation}
  where $m=0$ specify that we replace conditioning frames with zeros and $m=1$ otherwise.
  We employ context guidance \cite{blattmann2023align} at inference, with $w$ representing the guidance weight:

  \begin{equation}      
    \begin{split}
      \tilde{z}_{\theta, \phi}([z_t, c], tp, s) &= (1+w) z_{\theta, \phi}([z_t, c], tp, s, m=1)\\ &- w z_{\theta, \phi}([z_t, c], s, m=0)
    \end{split}
  \end{equation}

  Our interpolation model can benefit from a relatively low guidance weight value, such as $0.25$ or $0.5$. Increasing this value can significantly negatively impact the quality of the interpolated frames, sometimes resulting in the model generating frames that closely match the conditional keyframes (No actual interpolation).

\subsection{Video Decoder}
  
To enhance the video decoding process, we use a pretrained MoVQ-GAN \cite{zheng2022movq} model with a frozen encoder. 
  To extend the decoder into the temporal dimension, we explore several choices: Substituting 2D convolutions with 3D convolutions and adding temporal layers interleaved with existing spatial layers. Regarding the temporal layers, we explore the use of temporal convolutions, temporal 1D conv blocks, and temporal self-attentions. All additional parameters are initialized with zeros.

\section{Experiments}

\begin{figure*}[t]
        \center{\includegraphics[scale=0.16]{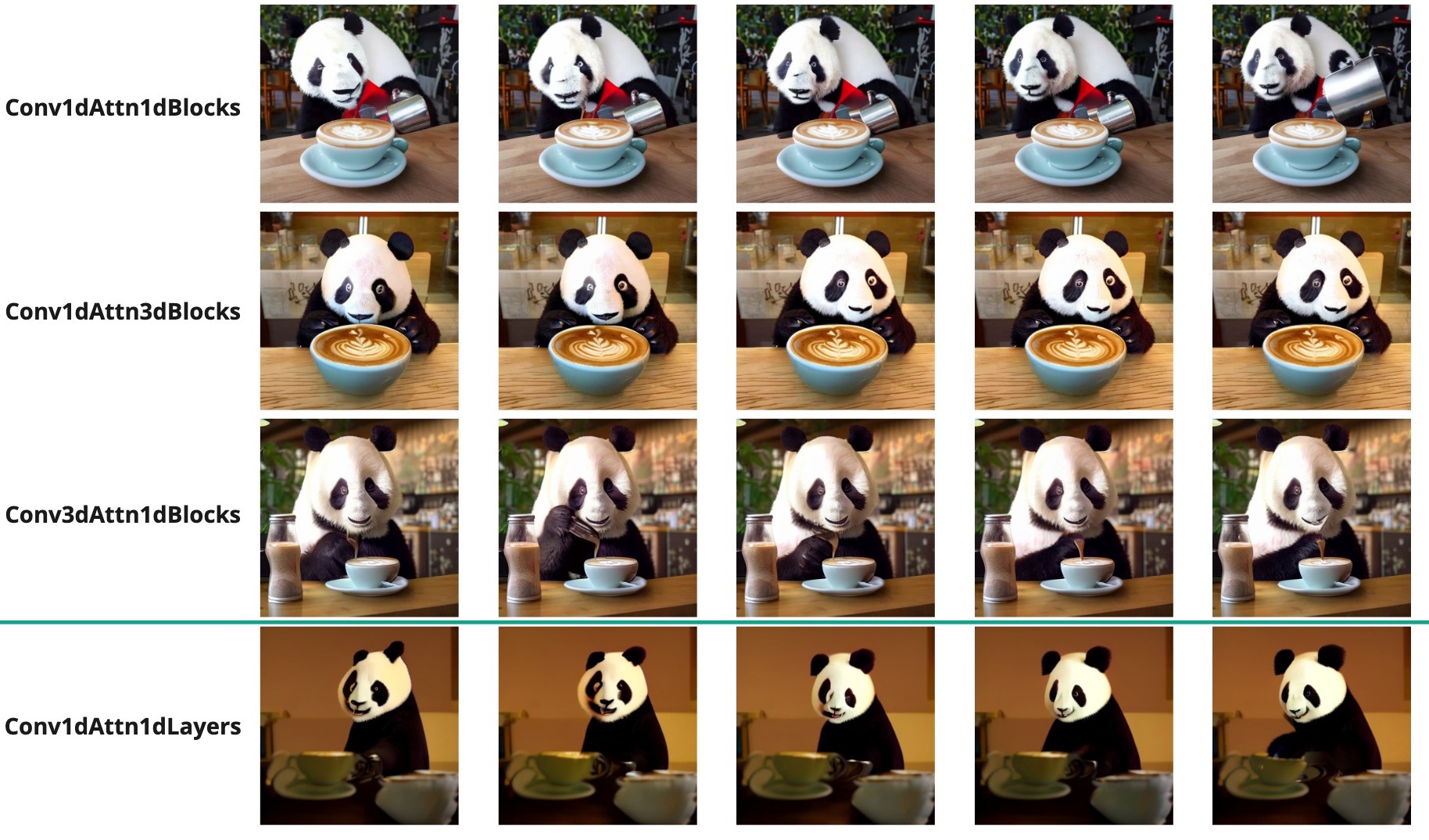}}
        \caption{Generations of keyframes for all our models. \textbf{Prompt}: \texttt{A panda making latte art in cafe.}}
       \label{fig:compare_1}
\end{figure*}

%\subsection{Experimental Setup}
    \label{subsec_experimental_setup}
    \textbf{Datasets.} Our internal training dataset for the keyframe generation model contains $120k$ text-video pairs, and the same dataset is utilized for training the interpolation model.
    In our evaluation, we evaluate the T2V model on two test sets: UCF-101 \cite{soomro2012ucf101} and MSR-VTT \cite{xu2016msr}.
    For training the decoder, we use a mix of $80k$ videos from the internal dataset, while model testing is performed on the septuplet part of Vimeo90k \cite{xue2019video} dataset.
    
    In preparing frame sequences for the interpolation task training, we randomly select a skip-frame value $s \in \{1, 2, \ldots, 12\}$. Subsequently, the input video is resampled with a skip of $s$ frames between each pair. This $33$ frames video is then organized into $8$ conditioning frames for each side and $8\times3$ target target frames. For decoder training, sequences comprising $8$ frames are employed.
        
    \textbf{Metrics.} In line with previous works \cite{singer2022make, luo2023videofusion, li2023videogen}, we assess our T2V model using the following evaluation metrics: Frechet Video Distance (FVD) \cite{unterthiner2018towards}, Inception Score (IS) \cite{saito2020train} and CLIPSIM \cite{wu2021godiva}. FVD assists in evaluating the fidelity of the generated video. IS metric assesses the quality and diversity of individual frames. CLIPSIM, on the other hand, evaluates text-video alignment. For FVD calculation, we use the protocol proposed in StyleGAN-V\cite{skorokhodov2022stylegan} for a fair comparison. For IS, there is no clear way to calculate in related works. In this work, we calculate IS on $2048$ videos by taking the first 16 frames in each video.
    For decoder training, other metrics are used: PSNR for frame quality assessment, SSIM for structural similarity evaluation, and LPIPS
    \cite{zhang2018unreasonable} for perceptual similarity. 
        
    \textbf{Training.}
    We trained 4 keyframe generation models for $100k$ steps on $8$ GPUs (A100, 80GB) with batch size of $1$, and gradient accumulation of $2$ to generate $16$ frames with $512 \times 512$ resolution. Keyframes generation was trained with $2$ FPS. As far as the FPS of training videos is around $24$, we sampled frames for training randomly choosing the first frame position and the positions of the next frames with the skip of $12$. Positions of sampled frames were encoded by a table of positional embeddings. We trained only temporal layers or temporal blocks, depending on the method. All other weights were taken from our T2I model and frozen. Parameters were optimized using AdamW optimizer with the constant learning rate of $0.0001$. The optimizer was patitioned across GPUs using ZeRO stage 1.
    
    The entire interpolation model was trained for $50k$ steps at the task of upsampling $9$ frames across different skip-frame values $s$ to a sequence of $33$ frames, all while maintaining a resolution of $256\times256$. During training, we set the probability for unconditional frame generation $uncond\_prob$ to $10\%$. Our decoder, including the spatial layers, is trained using sequences of $8$ frames for $50k$ steps. During interpolation training, we use $8$ GPUs (A100 80GB) and gradient accumulation steps of $4$. For decoder training, we turn off gradient accumulation.
    
    \textbf{Inference.} Keyframes are generated in the T2V generation phase. To interpolate between these generated keyframes, we employ the generated latents from the first phase as conditions to generate three middle frames between each keyframe pair. We set the skip-frame value $s$ to $3$ during the first step of interpolation (2FPS$\rightarrow$8FPS) and $1$ during the second step (8FPS$\rightarrow$30FPS). Additionally, we maintain a constant perturbation noise level $tp=0$. We set the guidance weight to a small value $w=0.25$. In the final stage, our trained decoder decodes the resulting latents and the latents from the generated keyframes to produce the final video output. 

\begin{table*}
  \centering
  \small
  \begin{tabular}{lcccc}
    \toprule
    Method & Zero-Shot & IS$\uparrow$ &  FVD$\downarrow$ &  CLIPSIM$\uparrow$\\
    \midrule
        \multicolumn{5}{c}{\textit{Proprietary Technologies}} \\
    \hline
    GoDIVA \cite{wu2021godiva} & No  & - & - & 0.2402 \\
    Nuwa \cite{wu2021nuwa} & No & - & - & 0.2439 \\
    Magic Video \cite{zhou2022magicvideo} & No & - & 699.00 & - \\
    Video LDM \cite{blattmann2023align} & No & - & 550.61 & 0.2929\\
    Make-A-Video \cite{singer2022make} & Yes & \textbf{33.00}  & \textbf{367.23} & \textbf{0.3049}\\
    \hline
        \multicolumn{5}{c}{\textit{Open Sourced Techologies}} \\
    \hline
    LVDM \cite{he2022latent} & No & - & 641.80 & - \\
    ModelScope \cite{wang2023modelscope} & No & - & - & 0.2930\\
    LaVie \cite{wang2023lavie}& No & - & 526.30 & 0.2949\\
    CogVideo (Chinese) \cite{hong2022cogvideo} & Yes & 23.55 &  751.34 & 0.2614\\
    CogVideo (English) \cite{hong2022cogvideo} & Yes & {25.27} & 710.59 & 0.2631\\
    \hline
    \textbf{Temporal Blocks (ours)}\\
    Conv1dAttn1dLayers (100k)& Yes & 19.663 & 659.612 & 0.2827\\
    Conv1dAttn1dBlocks (100k)& Yes & 23.063 & 545.184 & 0.2955\\
    Conv1dAttn3dBlocks (100k)& Yes & 23.381 & 573.569 & 0.2956\\
    Conv3dAttn1dBlocks (100k)& Yes & 22.899 & 594.919 & 0.2953\\
    Conv1dAttn1dBlocks (220k)& Yes & \textbf{24.325} & \textbf{433.054} & \textbf{0.2976}\\
    \bottomrule
  \end{tabular}
    \caption{\textbf{T2V results on UCF-101 and MSR-VTT.} We compare various options to build temporal blocks: 1) Conv1dAttn1dLayers, 2) Conv1dAttn1dBlocks, 3) Conv1dAttn3dBlocks, and 4) Conv1dAttn1dBlocks, (Between parentheses -- the number of training steps). We calculate FVD, IS on UCF-101 and CLIPSIM on MSR-VTT.}
    \label{table-t2v-ucf}
\end{table*}

\section{Results}
  \subsection{Quantitative Results}
    In this section, we provide a comparison of our trained models using FVD, IS on UCF-101 and CLIPSIM on MSR-VTT, as detailed in Table \ref{table-t2v-ucf}.
    First, we assess the models trained for $100k$ training steps. Concerning the comparison with \emph{Conv1dAttn1dLayers}, our results clearly indicate that the inclusion of temporal blocks, rather than temporal layers, results in significantly enhanced quality according to these metrics. 
    In the assessment of temporal blocks, \emph{Conv3dAttn1dBlocks} falls behind, achieving values of $594.919$ for FVD and $22.899$ for IS. 
    \emph{Conv3dAttn1dBlocks} shows an IS score of $23.381$ for IS  and an FVD score of $573.569$. 
    Comparing with \emph{Conv3dAttn1dBlocks}, \emph{Conv3dAttn1dBlocks} records a worse IS score of $23.063$ but a the best FVD score $545.184$.
    The CLIPSIM results do not show a clear difference among temporal blocks, with slightly better performance observed for \emph{Conv1dAttn3dBlocks}, reaching a CLIPSIM of $0.2956$. 
    
    Next, we showcase the best outcomes  achieved for \emph{Conv1dAttn1dBlocks} after being trained for $220k$ training steps. Evidently, our model experiencing significant improvements in terms of FVD ($433.504$) and IS ($24.325$).

    We want to highlight that the relatively lower performance in IS compared to the baselines could be attributed to ambiguities in IS assessment. Existing literature lacks sufficient details regarding the methodologies used for calculating IS on video in related works.
    
  \subsection{Qualitative Results}

    \begin{figure}[ht]
        \center{\includegraphics[scale=0.28]{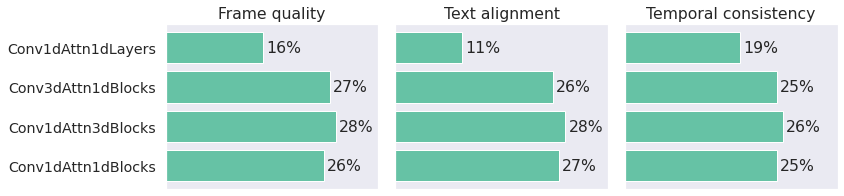}}
        \caption{\textbf{Human evaluation study results.}
        The bars in the plot correspond to the percentage of ``wins'' in the side-by-side comparison of model generations.}
        \label{fig:sbs}
    \end{figure}
  
    We performed a preference test to compare our models between each other. For this purpose, we created a bot that displays a pair of videos. An annotator then chooses the video with best quality in terms of three visual measurements: 1) frame quality, 2) alignment with text, and 3) temporal consistency. A total of $31$ annotators engaged in this process and the total number of used video pairs is $6600$. Figure \ref{fig:sbs} presents the results of this user study. There is a clear preference of videos generated by temporal blocks based models over the videos generated by temporal layer based model with a large margin among all measurements. 
    The temporal layer based model often produces either semantically unrelated keyframes, or does not deal well with the dynamics.
    Regarding the comparison between temporal block methods against each other, there is a small but consist preference for the model with Temp. Conv1D Block and Temp. Attn3D Block over the two other models. Finally, there is no clear preference between the rest two models. The qualitative comparison reveals visually observable advantages of our technique, both in the quality of generated objects on individual keyframes and in dynamics as illustrated in Figure \ref{fig:compare_1}. The method based on temporal blocks generates more consistent content and more coherent keyframes sequence. More examples with video generation can be found on the project page and in supplementary material.
    
    \subsection{Interpolation Architecture Evaluation}
\begin{figure*}[ht]
    \center{\includegraphics[scale=0.42]{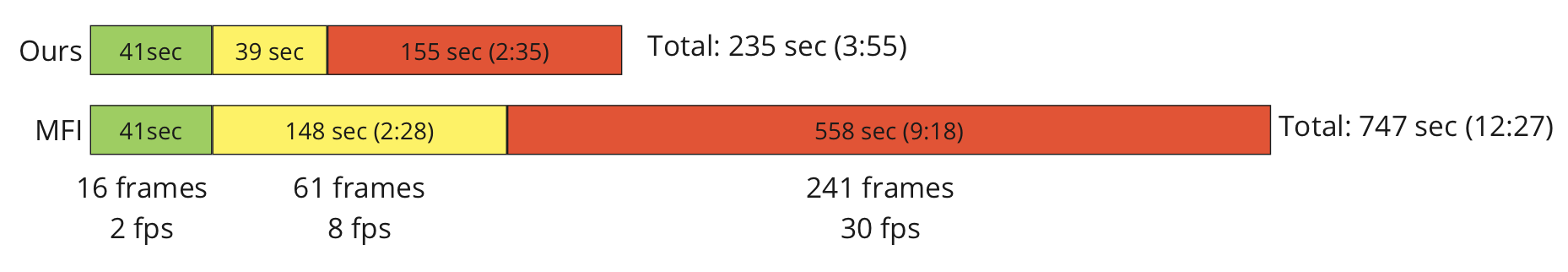}}
    \caption{\textbf{Running time comparison between two pipelines with different interpolation architectures:} (1) Our architecture, and (2) Masked Frame Interpolation (MFI) on A100 80GB. Keyframe generation in green. First and second step interpolation in yellow and red respectively.}
    \label{fig:interpolation_running}
\end{figure*}

    To assess the effectiveness of our interpolation architecture, we implemented Masked Frame Interpolation (MFI) architecture described in Video LDM \cite{blattmann2023align} and Appendix \ref{appendix_interpolation_modifications}. For our interpolation architecture, we use a $220k$ training steps checkpoint for a model trained with temporal 1D-Blocks. We generated interpolated frames for a set of $2048$ videos, all composed exclusively of generated keyframes. Results, as presented in Table~\ref{table-interp-temporal-masking}, reveal that our new interpolation architecture excels in generating interpolated frames with superior quality and higher fidelity. Specifically, it achieves FVD of $433.054$ compared to $550.932$ obtained with MFI. 
    The inference time associated with each architecture is illustrated in Figure~\ref{fig:interpolation_running}, demonstrating that our new architecture is more efficient, reducing the running time to less than third compared to MFI. 

\begin{table}
  \centering
  \small
  \begin{tabular}{lll}
    \toprule
    Method & IS$\uparrow$ & FVD$\downarrow$\\
    \midrule
    Masked Frame Interpolation  & 23.371 & 550.932 \\
    FusionFrames Interpolation & 24.325 & 433.054 \\
    \bottomrule
  \end{tabular}
\caption{A comparison on UCF-101 between our implementation of masked frame interpolation architecture and our new interpolation architecture after 2 interpolation steps (30 FPS).}   
\label{table-interp-temporal-masking}
\end{table}

\begin{table*}
  \centering
  \small
  \begin{tabular}{llllllll}
    \toprule
    Decoder & Temporal Layers &Finetune&PSNR$\uparrow$&SSIM$\uparrow$&MSE$\downarrow$&LPIPS$\downarrow$&\# Params\\
    \midrule
    Image & - & - & 32.9677&0.9056&0.0008&0.0049 & 161 M\\
    Video & 3x1x1 Conv & Temporal & 32.2544&0.893&0.0009&0.006 & 203 M\\
    Video & 3x3x3 Conv & Temporal & 33.5819&0.9111&0.0007&0.0044 & 539 M\\
    Video & 3x1x1 Conv & Decoder & 33.5051&0.9106&0.0007&0.0044 & 203 M\\
    Video & 3x3x3 Conv & Decoder & 33.6342&0.9123&0.0007&0.0043 & 539 M\\
    Video & 3x1x1 Conv + Attn & Decoder & 33.7343&0.9129&0.0007&0.0043 & 220 M\\
    Video & 3x3x3 Conv + Attn & Decoder & \textbf{33.8376}&\textbf{0.9146}&\textbf{0.0006}&\textbf{0.0041} & 556 M\\
    Video & ResNet Block + Attn & Decoder & 33.7024&0.9121&0.0007&0.0043 & 220 M\\
    Video & 2D $\rightarrow$ 3D Conv & Decoder & 33.7321&0.9134&0.0007&0.0043 & 419 M \\
    \bottomrule
  \end{tabular}
  \caption{\textbf{A comparison between different choices to construct video decoder.} Including the use of temporal convolution, temporal ResNet Block, temporal attention (Attn) and finally converting 2D spatial convolution in the decoder into 3D conv (2D$\rightarrow$3D Conv). We also present whether we only finetune temporal layers or the entire decoder.}
    \label{table-video-movqgan}
\end{table*}

\subsection{Video MoVQ-GAN Experiments}\label{sec:ablation_study}
  We conducted comprehensive experiments, considering many choices of how to build video decoder, and assessed them in terms of quality metrics and the impact on the number of additional parameters. The results are presented in Table~\ref{table-video-movqgan}. This evaluation process guided us in making the optimal choice for production purposes. Extending the decoder with a $3\times3\times3$ temporal convolution and incorporating temporal attention during the fine-tuning process yields the highest overall quality among the available options. In this manner, we apply finetuning to the entire decoder, including spatial layers and the newly introduced parameters.  An alternative efficient choice involves using a $3\times1\times1$ temporal convolution layer or temporal 1D Block with temporal attention, which significantly reduces the number of parameters from $556M$  to $220M$ while still achieving results that closely match the quality obtained through the more extensive approach.

\section{Limitations}
  Comparative analysis faces challenges due to the ambiguities in the calculation procedures of metrics such as FVD and IS in related works (Also, noted by \cite{skorokhodov2022stylegan}), and that hiders the ability to draw meaningful comparisons with other studies. We recommend to use the protocol for FVD calculation described in StyleGAN-V\cite{skorokhodov2022stylegan}. Metrics obtained using this protocol (and their evaluation code) closely match the measurements of our own implementation.
  Furthermore, comparing our interpolation network with existing works poses challenges, and the absence of open solutions for interpolation in the latent space forces us to independently implement the other described approach~\cite{blattmann2023align}.
\section{Conclusion}

In this research, we examined several aspects of T2V architecture design in order to get the most considerable output quality. This challenging task included the development of a two-stage model for video synthesis, considering several ways to incorporate temporal information: temporal blocks and temporal layers. According to experiments, the first approach leads to higher metrics values in terms of visual quality measured by IS and FVD scores. We achieved a comparable IS score value to several existing solutions and a top-2 scores overall and top-1 among open-source T2V models in terms of CLIPSIM and FVD metrics. The interpolation architecture presented in this work excels in generating high-quality interpolated frames and surpasses masked frame interpolation architecture in terms of quality measured by IS and FVD. Our interpolation architecture is also more efficient in execution time and its execution time is more than 3 times effective comparing to the well-known masked frame interpolation approach. The paper also presents a new MoVQ-based video decoding scheme and the results of its effect on the overall quality. The following problems still remain and need to be researched further: the image quality of frames, adjacent frame  smoothness consistency with visual quality preserving.

{
    \small
    \bibliographystyle{ieeenat_fullname}
    \bibliography{main}

\begin{thebibliography}{54}
\providecommand{\natexlab}[1]{#1}
\providecommand{\url}[1]{\texttt{#1}}
\expandafter\ifx\csname urlstyle\endcsname\relax
  \providecommand{\doi}[1]{doi: #1}\else
  \providecommand{\doi}{doi: \begingroup \urlstyle{rm}\Url}\fi

\bibitem[Arkhipkin et~al.(2023)Arkhipkin, Filatov, Vasilev, Maltseva, Azizov, Pavlov, Agafonova, Kuznetsov, and Dimitrov]{arkhipkin2023kandinsky}
Vladimir Arkhipkin, Andrei Filatov, Viacheslav Vasilev, Anastasia Maltseva, Said Azizov, Igor Pavlov, Julia Agafonova, Andrey Kuznetsov, and Denis Dimitrov.
\newblock Kandinsky 3.0 technical report, 2023.

\bibitem[Babaeizadeh et~al.(2017)Babaeizadeh, Finn, Erhan, Campbell, and Levine]{babaeizadeh2017stochastic}
Mohammad Babaeizadeh, Chelsea Finn, Dumitru Erhan, Roy~H Campbell, and Sergey Levine.
\newblock Stochastic variational video prediction.
\newblock \emph{arXiv preprint arXiv:1710.11252}, 2017.

\bibitem[Babaeizadeh et~al.(2021)Babaeizadeh, Saffar, Nair, Levine, Finn, and Erhan]{babaeizadeh2021fitvid}
Mohammad Babaeizadeh, Mohammad~Taghi Saffar, Suraj Nair, Sergey Levine, Chelsea Finn, and Dumitru Erhan.
\newblock Fitvid: Overfitting in pixel-level video prediction.
\newblock \emph{arXiv preprint arXiv:2106.13195}, 2021.

\bibitem[Blattmann et~al.(2023)Blattmann, Rombach, Ling, Dockhorn, Kim, Fidler, and Kreis]{blattmann2023align}
Andreas Blattmann, Robin Rombach, Huan Ling, Tim Dockhorn, Seung~Wook Kim, Sanja Fidler, and Karsten Kreis.
\newblock Align your latents: High-resolution video synthesis with latent diffusion models.
\newblock In \emph{Proceedings of the IEEE/CVF Conference on Computer Vision and Pattern Recognition}, pages 22563--22575, 2023.

\bibitem[Clark et~al.(2019)Clark, Donahue, and Simonyan]{clark2019adversarial}
Aidan Clark, Jeff Donahue, and Karen Simonyan.
\newblock Adversarial video generation on complex datasets.
\newblock \emph{arXiv preprint arXiv:1907.06571}, 2019.

\bibitem[Ding et~al.(2022)Ding, Zheng, Hong, and Tang]{ding2022cogview2}
Ming Ding, Wendi Zheng, Wenyi Hong, and Jie Tang.
\newblock Cogview2: Faster and better text-to-image generation via hierarchical transformers.
\newblock \emph{Advances in Neural Information Processing Systems}, 35:\penalty0 16890--16902, 2022.

\bibitem[Esser et~al.(2023)Esser, Chiu, Atighehchian, Granskog, and Germanidis]{esser2023gen1}
Patrick Esser, Johnathan Chiu, Parmida Atighehchian, Jonathan Granskog, and Anastasis Germanidis.
\newblock Structure and content-guided video synthesis with diffusion models, 2023.

\bibitem[Ge et~al.(2022)Ge, Hayes, Yang, Yin, Pang, Jacobs, Huang, and Parikh]{ge2022long}
Songwei Ge, Thomas Hayes, Harry Yang, Xi Yin, Guan Pang, David Jacobs, Jia-Bin Huang, and Devi Parikh.
\newblock Long video generation with time-agnostic vqgan and time-sensitive transformer.
\newblock \emph{arXiv preprint arXiv:2204.03638}, 2022.

\bibitem[He et~al.(2022)He, Yang, Zhang, Shan, and Chen]{he2022latent}
Yingqing He, Tianyu Yang, Yong Zhang, Ying Shan, and Qifeng Chen.
\newblock Latent video diffusion models for high-fidelity video generation with arbitrary lengths.
\newblock \emph{arXiv preprint arXiv:2211.13221}, 2022.

\bibitem[Ho et~al.(2020)Ho, Jain, and Abbeel]{ho2020denoising}
Jonathan Ho, Ajay Jain, and Pieter Abbeel.
\newblock Denoising diffusion probabilistic models.
\newblock \emph{Advances in neural information processing systems}, 33:\penalty0 6840--6851, 2020.

\bibitem[Ho et~al.(2022{\natexlab{a}})Ho, Chan, Saharia, Whang, Gao, Gritsenko, Kingma, Poole, Norouzi, Fleet, et~al.]{ho2022imagen}
Jonathan Ho, William Chan, Chitwan Saharia, Jay Whang, Ruiqi Gao, Alexey Gritsenko, Diederik~P Kingma, Ben Poole, Mohammad Norouzi, David~J Fleet, et~al.
\newblock Imagen video: High definition video generation with diffusion models.
\newblock \emph{arXiv preprint arXiv:2210.02303}, 2022{\natexlab{a}}.

\bibitem[Ho et~al.(2022{\natexlab{b}})Ho, Salimans, Gritsenko, Chan, Norouzi, and Fleet]{ho2022video}
Jonathan Ho, Tim Salimans, Alexey Gritsenko, William Chan, Mohammad Norouzi, and David~J Fleet.
\newblock Video diffusion models.
\newblock \emph{arXiv:2204.03458}, 2022{\natexlab{b}}.

\bibitem[Hong et~al.(2022)Hong, Ding, Zheng, Liu, and Tang]{hong2022cogvideo}
Wenyi Hong, Ming Ding, Wendi Zheng, Xinghan Liu, and Jie Tang.
\newblock Cogvideo: Large-scale pretraining for text-to-video generation via transformers.
\newblock \emph{arXiv preprint arXiv:2205.15868}, 2022.

\bibitem[Kong and Ping(2021)]{kong2021fast}
Zhifeng Kong and Wei Ping.
\newblock On fast sampling of diffusion probabilistic models.
\newblock \emph{arXiv preprint arXiv:2106.00132}, 2021.

\bibitem[Kumar et~al.(2019)Kumar, Babaeizadeh, Erhan, Finn, Levine, Dinh, and Kingma]{kumar2019videoflow}
Manoj Kumar, Mohammad Babaeizadeh, Dumitru Erhan, Chelsea Finn, Sergey Levine, Laurent Dinh, and Durk Kingma.
\newblock Videoflow: A flow-based generative model for video.
\newblock \emph{arXiv preprint arXiv:1903.01434}, 2\penalty0 (5):\penalty0 3, 2019.

\bibitem[Lee et~al.(2018)Lee, Zhang, Ebert, Abbeel, Finn, and Levine]{lee2018stochastic}
Alex~X Lee, Richard Zhang, Frederik Ebert, Pieter Abbeel, Chelsea Finn, and Sergey Levine.
\newblock Stochastic adversarial video prediction.
\newblock \emph{arXiv preprint arXiv:1804.01523}, 2018.

\bibitem[Li et~al.(2023)Li, Chu, Wu, Yuan, Liu, Zhang, Li, Feng, Ding, and Wang]{li2023videogen}
Xin Li, Wenqing Chu, Ye Wu, Weihang Yuan, Fanglong Liu, Qi Zhang, Fu Li, Haocheng Feng, Errui Ding, and Jingdong Wang.
\newblock Videogen: A reference-guided latent diffusion approach for high definition text-to-video generation.
\newblock \emph{arXiv preprint arXiv:2309.00398}, 2023.

\bibitem[Li et~al.(2018)Li, Min, Shen, Carlson, and Carin]{li2018video}
Yitong Li, Martin Min, Dinghan Shen, David Carlson, and Lawrence Carin.
\newblock Video generation from text.
\newblock In \emph{Proceedings of the AAAI conference on artificial intelligence}, 2018.

\bibitem[Luo et~al.(2023)Luo, Chen, Zhang, Huang, Wang, Shen, Zhao, Zhou, and Tan]{luo2023videofusion}
Zhengxiong Luo, Dayou Chen, Yingya Zhang, Yan Huang, Liang Wang, Yujun Shen, Deli Zhao, Jingren Zhou, and Tieniu Tan.
\newblock Videofusion: Decomposed diffusion models for high-quality video generation.
\newblock In \emph{Proceedings of the IEEE/CVF Conference on Computer Vision and Pattern Recognition}, pages 10209--10218, 2023.

\bibitem[Mittal et~al.(2017)Mittal, Marwah, and Balasubramanian]{mittal2017sync}
Gaurav Mittal, Tanya Marwah, and Vineeth~N Balasubramanian.
\newblock Sync-draw: Automatic video generation using deep recurrent attentive architectures.
\newblock In \emph{Proceedings of the 25th ACM international conference on Multimedia}, pages 1096--1104, 2017.

\bibitem[Ni et~al.(2023)Ni, Shi, Li, Huang, and Min]{ni2023conditional}
Haomiao Ni, Changhao Shi, Kai Li, Sharon~X Huang, and Martin~Renqiang Min.
\newblock Conditional image-to-video generation with latent flow diffusion models.
\newblock In \emph{Proceedings of the IEEE/CVF Conference on Computer Vision and Pattern Recognition}, pages 18444--18455, 2023.

\bibitem[Nichol et~al.(2022)Nichol, Dhariwal, Ramesh, Shyam, Mishkin, McGrew, Sutskever, and Chen]{Nichol2022glide}
Alexander~Quinn Nichol, Prafulla Dhariwal, Aditya Ramesh, Pranav Shyam, Pamela Mishkin, Bob McGrew, Ilya Sutskever, and Mark Chen.
\newblock {GLIDE:} towards photorealistic image generation and editing with text-guided diffusion models.
\newblock In \emph{International Conference on Machine Learning, {ICML} 2022, 17-23 July 2022, Baltimore, Maryland, {USA}}, pages 16784--16804. {PMLR}, 2022.

\bibitem[Pan et~al.(2017)Pan, Qiu, Yao, Li, and Mei]{pan2017create}
Yingwei Pan, Zhaofan Qiu, Ting Yao, Houqiang Li, and Tao Mei.
\newblock To create what you tell: Generating videos from captions.
\newblock In \emph{Proceedings of the 25th ACM international conference on Multimedia}, pages 1789--1798, 2017.

\bibitem[Ramesh et~al.(2022)Ramesh, Dhariwal, Nichol, Chu, and Chen]{ramesh2022hierarchical}
Aditya Ramesh, Prafulla Dhariwal, Alex Nichol, Casey Chu, and Mark Chen.
\newblock Hierarchical text-conditional image generation with clip latents.
\newblock \emph{arXiv preprint arXiv:2204.06125}, 1\penalty0 (2):\penalty0 3, 2022.

\bibitem[Rombach et~al.(2022)Rombach, Blattmann, Lorenz, Esser, and Ommer]{rombach2022high}
Robin Rombach, Andreas Blattmann, Dominik Lorenz, Patrick Esser, and Bj{\"o}rn Ommer.
\newblock High-resolution image synthesis with latent diffusion models.
\newblock In \emph{Proceedings of the IEEE/CVF conference on computer vision and pattern recognition}, pages 10684--10695, 2022.

\bibitem[Saharia et~al.(2022)Saharia, Chan, Saxena, Li, Whang, Denton, Ghasemipour, Gontijo~Lopes, Karagol~Ayan, Salimans, et~al.]{saharia2022photorealistic}
Chitwan Saharia, William Chan, Saurabh Saxena, Lala Li, Jay Whang, Emily~L Denton, Kamyar Ghasemipour, Raphael Gontijo~Lopes, Burcu Karagol~Ayan, Tim Salimans, et~al.
\newblock Photorealistic text-to-image diffusion models with deep language understanding.
\newblock \emph{Advances in Neural Information Processing Systems}, 35:\penalty0 36479--36494, 2022.

\bibitem[Saito et~al.(2020)Saito, Saito, Koyama, and Kobayashi]{saito2020train}
Masaki Saito, Shunta Saito, Masanori Koyama, and Sosuke Kobayashi.
\newblock Train sparsely, generate densely: Memory-efficient unsupervised training of high-resolution temporal gan.
\newblock \emph{International Journal of Computer Vision}, 128\penalty0 (10-11):\penalty0 2586--2606, 2020.

\bibitem[Salimans and Ho(2022)]{salimans2022progressive}
Tim Salimans and Jonathan Ho.
\newblock Progressive distillation for fast sampling of diffusion models.
\newblock \emph{arXiv preprint arXiv:2202.00512}, 2022.

\bibitem[Singer et~al.(2022)Singer, Polyak, Hayes, Yin, An, Zhang, Hu, Yang, Ashual, Gafni, et~al.]{singer2022make}
Uriel Singer, Adam Polyak, Thomas Hayes, Xi Yin, Jie An, Songyang Zhang, Qiyuan Hu, Harry Yang, Oron Ashual, Oran Gafni, et~al.
\newblock Make-a-video: Text-to-video generation without text-video data.
\newblock \emph{arXiv preprint arXiv:2209.14792}, 2022.

\bibitem[Skorokhodov et~al.(2022)Skorokhodov, Tulyakov, and Elhoseiny]{skorokhodov2022stylegan}
Ivan Skorokhodov, Sergey Tulyakov, and Mohamed Elhoseiny.
\newblock Stylegan-v: A continuous video generator with the price, image quality and perks of stylegan2.
\newblock In \emph{Proceedings of the IEEE/CVF Conference on Computer Vision and Pattern Recognition}, pages 3626--3636, 2022.

\bibitem[Sohl-Dickstein et~al.(2015)Sohl-Dickstein, Weiss, Maheswaranathan, and Ganguli]{Dickstein2015}
Jascha Sohl-Dickstein, Eric Weiss, Niru Maheswaranathan, and Surya Ganguli.
\newblock Deep unsupervised learning using nonequilibrium thermodynamics.
\newblock In \emph{Proceedings of the 32nd International Conference on Machine Learning}, pages 2256--2265, Lille, France, 2015. PMLR.

\bibitem[Song et~al.(2021)Song, Sohl{-}Dickstein, Kingma, Kumar, Ermon, and Poole]{Song2021SBD}
Yang Song, Jascha Sohl{-}Dickstein, Diederik~P. Kingma, Abhishek Kumar, Stefano Ermon, and Ben Poole.
\newblock Score-based generative modeling through stochastic differential equations.
\newblock In \emph{9th International Conference on Learning Representations, {ICLR} 2021, Virtual Event, Austria, May 3-7, 2021}. OpenReview.net, 2021.

\bibitem[Soomro et~al.(2012)Soomro, Zamir, and Shah]{soomro2012ucf101}
Khurram Soomro, Amir~Roshan Zamir, and Mubarak Shah.
\newblock Ucf101: A dataset of 101 human actions classes from videos in the wild.
\newblock \emph{arXiv preprint arXiv:1212.0402}, 2012.

\bibitem[Tu et~al.(2022)Tu, Talebi, Zhang, Yang, Milanfar, Bovik, and Li]{tu2022maxvit}
Zhengzhong Tu, Hossein Talebi, Han Zhang, Feng Yang, Peyman Milanfar, Alan Bovik, and Yinxiao Li.
\newblock Maxvit: Multi-axis vision transformer.
\newblock \emph{ECCV}, 2022.

\bibitem[Unterthiner et~al.(2018)Unterthiner, Van~Steenkiste, Kurach, Marinier, Michalski, and Gelly]{unterthiner2018towards}
Thomas Unterthiner, Sjoerd Van~Steenkiste, Karol Kurach, Raphael Marinier, Marcin Michalski, and Sylvain Gelly.
\newblock Towards accurate generative models of video: A new metric \& challenges.
\newblock \emph{arXiv preprint arXiv:1812.01717}, 2018.

\bibitem[Villegas et~al.(2022)Villegas, Babaeizadeh, Kindermans, Moraldo, Zhang, Saffar, Castro, Kunze, and Erhan]{Villegas2022Phenaki}
Ruben Villegas, Mohammad Babaeizadeh, Pieter-Jan Kindermans, Hernan Moraldo, Han Zhang, Mohammad~Taghi Saffar, Santiago Castro, Julius Kunze, and D. Erhan.
\newblock Phenaki: Variable length video generation from open domain textual description.
\newblock \emph{ArXiv}, abs/2210.02399, 2022.

\bibitem[Voleti et~al.(2022)Voleti, Jolicoeur-Martineau, and Pal]{voleti2022mcvd}
Vikram Voleti, Alexia Jolicoeur-Martineau, and Chris Pal.
\newblock Mcvd-masked conditional video diffusion for prediction, generation, and interpolation.
\newblock \emph{Advances in Neural Information Processing Systems}, 35:\penalty0 23371--23385, 2022.

\bibitem[Vondrick et~al.(2016)Vondrick, Pirsiavash, and Torralba]{Vondrick2016Videos}
Carl Vondrick, Hamed Pirsiavash, and Antonio Torralba.
\newblock Generating videos with scene dynamics.
\newblock In \emph{Proceedings of the 30th International Conference on Neural Information Processing Systems}, page 613–621, Red Hook, NY, USA, 2016. Curran Associates Inc.

\bibitem[Walker et~al.(2021)Walker, Razavi, and Oord]{walker2021predicting}
Jacob Walker, Ali Razavi, and A{\"a}ron van~den Oord.
\newblock Predicting video with vqvae.
\newblock \emph{arXiv preprint arXiv:2103.01950}, 2021.

\bibitem[Wang et~al.(2023{\natexlab{a}})Wang, Yuan, Chen, Zhang, Wang, and Zhang]{wang2023modelscope}
Jiuniu Wang, Hangjie Yuan, Dayou Chen, Yingya Zhang, Xiang Wang, and Shiwei Zhang.
\newblock Modelscope text-to-video technical report, 2023{\natexlab{a}}.

\bibitem[Wang et~al.(2023{\natexlab{b}})Wang, Chen, Ma, Zhou, Huang, Wang, Yang, He, Yu, Yang, Guo, Wu, Si, Jiang, Chen, Loy, Dai, Lin, Qiao, and Liu]{wang2023lavie}
Yaohui Wang, Xinyuan Chen, Xin Ma, Shangchen Zhou, Ziqi Huang, Yi Wang, Ceyuan Yang, Yinan He, Jiashuo Yu, Peiqing Yang, Yuwei Guo, Tianxing Wu, Chenyang Si, Yuming Jiang, Cunjian Chen, Chen~Change Loy, Bo Dai, Dahua Lin, Yu Qiao, and Ziwei Liu.
\newblock Lavie: High-quality video generation with cascaded latent diffusion models, 2023{\natexlab{b}}.

\bibitem[Wu et~al.(2021{\natexlab{a}})Wu, Huang, Zhang, Li, Ji, Yang, Sapiro, and Duan]{wu2021godiva}
Chenfei Wu, Lun Huang, Qianxi Zhang, Binyang Li, Lei Ji, Fan Yang, Guillermo Sapiro, and Nan Duan.
\newblock Godiva: Generating open-domain videos from natural descriptions.
\newblock \emph{arXiv preprint arXiv:2104.14806}, 2021{\natexlab{a}}.

\bibitem[Wu et~al.(2021{\natexlab{b}})Wu, Liang, Ji, Yang, Fang, Jiang, and Duan]{wu2021nuwa}
Chenfei Wu, Jian Liang, Lei Ji, Fan Yang, Yuejian Fang, Daxin Jiang, and Nan Duan.
\newblock N\"uwa: Visual synthesis pre-training for neural visual world creation, 2021{\natexlab{b}}.

\bibitem[Wu et~al.(2022)Wu, Ge, Wang, Lei, Gu, Hsu, Shan, Qie, and Shou]{wu2022tuneavideo}
Jay~Zhangjie Wu, Yixiao Ge, Xintao Wang, Stan~Weixian Lei, Yuchao Gu, Wynne Hsu, Ying Shan, Xiaohu Qie, and Mike~Zheng Shou.
\newblock Tune-a-video: One-shot tuning of image diffusion models for text-to-video generation.
\newblock \emph{arXiv preprint arXiv:2212.11565}, 2022.

\bibitem[Xu et~al.(2016)Xu, Mei, Yao, and Rui]{xu2016msr}
Jun Xu, Tao Mei, Ting Yao, and Yong Rui.
\newblock Msr-vtt: A large video description dataset for bridging video and language.
\newblock In \emph{Proceedings of the IEEE conference on computer vision and pattern recognition}, pages 5288--5296, 2016.

\bibitem[Xue et~al.(2019)Xue, Chen, Wu, Wei, and Freeman]{xue2019video}
Tianfan Xue, Baian Chen, Jiajun Wu, Donglai Wei, and William~T Freeman.
\newblock Video enhancement with task-oriented flow.
\newblock \emph{International Journal of Computer Vision}, 127:\penalty0 1106--1125, 2019.

\bibitem[Yan et~al.(2021)Yan, Zhang, Abbeel, and Srinivas]{yan2021videogpt}
Wilson Yan, Yunzhi Zhang, Pieter Abbeel, and Aravind Srinivas.
\newblock Videogpt: Video generation using vq-vae and transformers.
\newblock \emph{arXiv preprint arXiv:2104.10157}, 2021.

\bibitem[Yang et~al.(2022)Yang, Srivastava, and Mandt]{yang2022diffusion}
Ruihan Yang, Prakhar Srivastava, and Stephan Mandt.
\newblock Diffusion probabilistic modeling for video generation.
\newblock \emph{arXiv preprint arXiv:2203.09481}, 2022.

\bibitem[Yu et~al.(2023)Yu, Sohn, Kim, and Shin]{yu2023video}
Sihyun Yu, Kihyuk Sohn, Subin Kim, and Jinwoo Shin.
\newblock Video probabilistic diffusion models in projected latent space.
\newblock In \emph{Proceedings of the IEEE/CVF Conference on Computer Vision and Pattern Recognition}, 2023.

\bibitem[Zhang et~al.(2023{\natexlab{a}})Zhang, Wu, Liu, Zhao, Ran, Gu, Gao, and Shou]{zhang2023show}
David~Junhao Zhang, Jay~Zhangjie Wu, Jia-Wei Liu, Rui Zhao, Lingmin Ran, Yuchao Gu, Difei Gao, and Mike~Zheng Shou.
\newblock Show-1: Marrying pixel and latent diffusion models for text-to-video generation.
\newblock \emph{arXiv preprint arXiv:2309.15818}, 2023{\natexlab{a}}.

\bibitem[Zhang et~al.(2018)Zhang, Isola, Efros, Shechtman, and Wang]{zhang2018unreasonable}
Richard Zhang, Phillip Isola, Alexei~A Efros, Eli Shechtman, and Oliver Wang.
\newblock The unreasonable effectiveness of deep features as a perceptual metric.
\newblock In \emph{Proceedings of the IEEE conference on computer vision and pattern recognition}, pages 586--595, 2018.

\bibitem[Zhang et~al.(2023{\natexlab{b}})Zhang, Wei, Jiang, Zhang, Zuo, and Tian]{zhang2023controlvideo}
Yabo Zhang, Yuxiang Wei, Dongsheng Jiang, Xiaopeng Zhang, Wangmeng Zuo, and Qi Tian.
\newblock Controlvideo: Training-free controllable text-to-video generation.
\newblock \emph{arXiv preprint arXiv:2305.13077}, 2023{\natexlab{b}}.

\bibitem[Zheng et~al.(2022)Zheng, Vuong, Cai, and Phung]{zheng2022movq}
Chuanxia Zheng, Tung-Long Vuong, Jianfei Cai, and Dinh Phung.
\newblock Movq: Modulating quantized vectors for high-fidelity image generation.
\newblock \emph{Advances in Neural Information Processing Systems}, 35:\penalty0 23412--23425, 2022.

\bibitem[Zhou et~al.(2022)Zhou, Wang, Yan, Lv, Zhu, and Feng]{zhou2022magicvideo}
Daquan Zhou, Weimin Wang, Hanshu Yan, Weiwei Lv, Yizhe Zhu, and Jiashi Feng.
\newblock Magicvideo: Efficient video generation with latent diffusion models.
\newblock \emph{arXiv preprint arXiv:2211.11018}, 2022.

\end{thebibliography}
}

\clearpage
\setcounter{page}{1}
\maketitlesupplementary

\begin{figure*}[htb]
    \center{\includegraphics[scale=0.15]{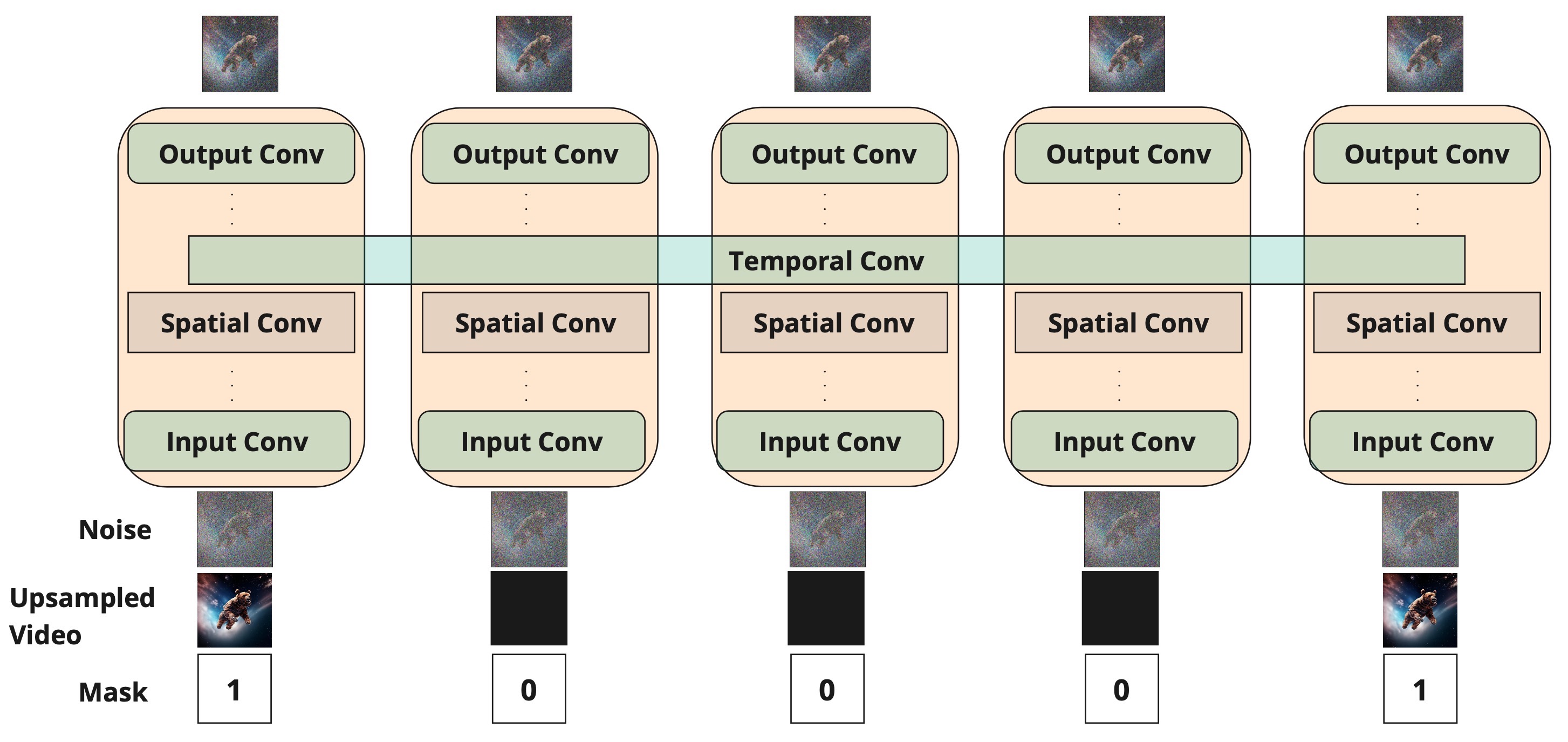}}
    \caption{\textbf{Masked Frame Interpolation Architecture \cite{blattmann2023align}.} Input to unit consists of noisy latent input, upsampled latent video and a mask. The mask specifies which latent inputs correspond to conditional keyframes.}
    \label{fig:interpolation_mfi}
\end{figure*}

\section{FusionFrames Interpolation Architecture Implementation Details} \label{appendix_interpolation_modifications}

  Expanding the Text-to-Image architecture to effectively handle and generate a sequence of interpolated frames requires a series of modifications designed to process the data across the temporal dimension. In addition to this, further adjustments are necessary to facilitate the generation of three middle frames between two keyframes. 
  
  Specifically, starting with pre-trained weights for the Text-to-Image model, we replicate the weights within the output convolution layer three times. This transformation alters the dimensions of those weights from $(out\_channels, input\_channels, 3,3)$ to $(3 * out\_channels, input\_channels, 3,3)$, and a similar modification is carried out for the bias parameters, shifting them from $(out\_channels)$ to $(3 * out\_channels)$. This adaptation enables the generation of three identical frames before the training phase starts. 
  In the input convolution layer, we make similar adjustments to the input channels, initializing additional weights as zeros. Subsequently, a temporal convolution layer, with a kernel size of $3\times1\times1$, is introduced after each spatial convolution layer. The output from each temporal layer is then combined with the output from the spatial layer using a learned merge parameter $\alpha$.

  The spatial layers are designed to handle input as a batch of individual frames. When dealing with video input, a necessary adjustment involves reshaping the data to shift the temporal axis into the batch dimension. Consequently, before forwarding the activations through the temporal layers, a transformation is performed to revert the activations back to its original video dimensions. 

\section{Masked Frame Interpolation Implementation}
\label{appendix_masked_frame_interpolation}

  We re-implemented Masked Frame Interpolation as described in Video LDM\cite{blattmann2023align} and Fig. \ref{fig:interpolation_mfi}. U-Net takes upsampled video as input in addition to noisy frames and mask. We obtain upsampled video by  zero-padding in the place of interpolated frames. The mask specify which frames are key-frames. Conditional
  frame perturbation is also incorporated as described. We train a model for two stages interpolation: $T\rightarrow4T$ and $4T\rightarrow16T$. We initialize this model with the same pre-trained T2I weights used in our experiments. Also, we initialize additional parameters with zeros. The training example consists of a sequence of $5$ frames ($2$ conditional frames and $3$ interpolated frames). We train on the same number of GPUs utilized in our experiments for $17k$ steps (this number of training steps is used in Video LDM, and we found the model to converge around this number of training steps).

\newpage

\section{Additional Generation Results}

\begin{figure*}[ht]
    \center{\includegraphics[scale=0.44]{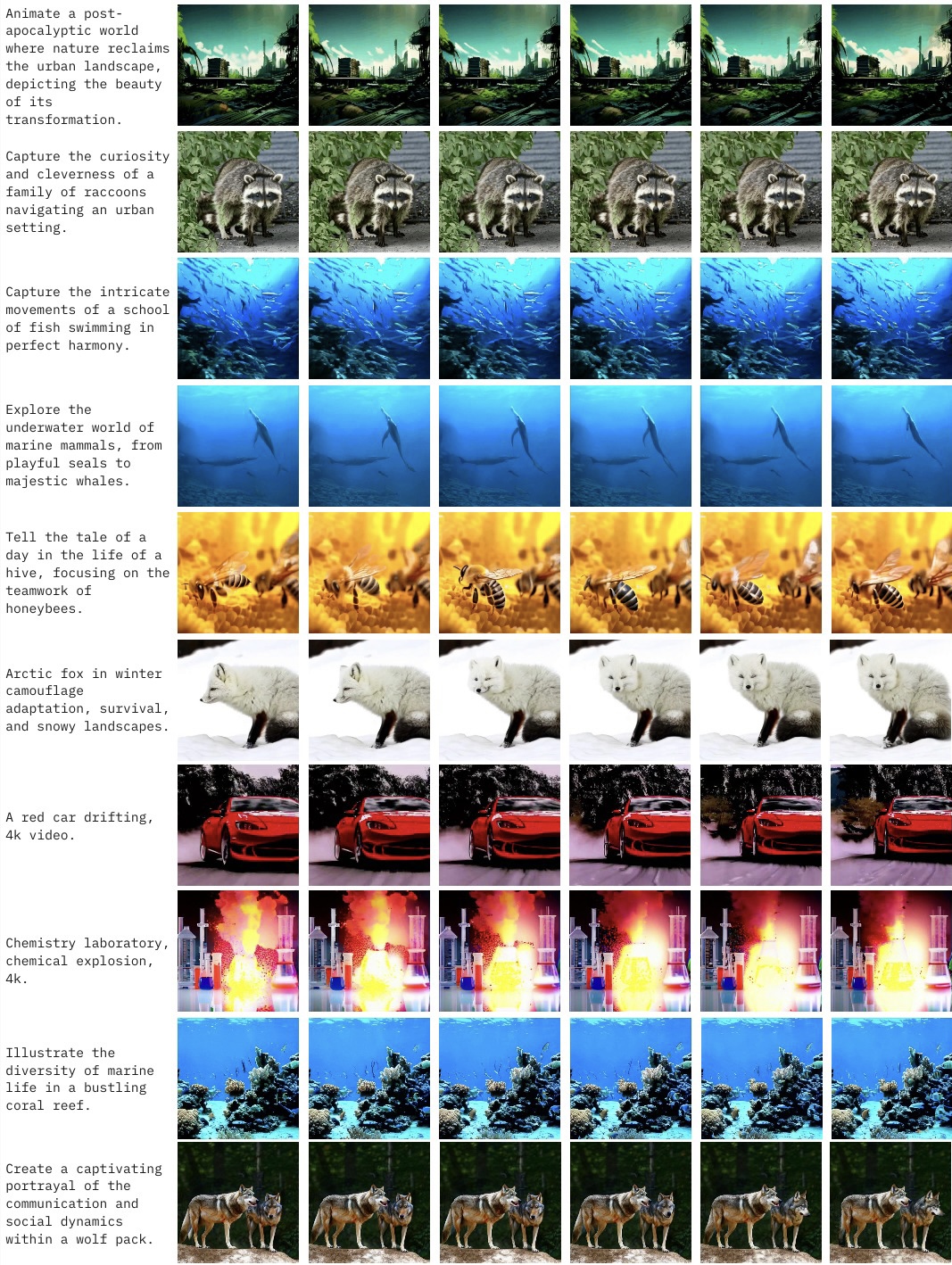}}
    \caption{\textbf{Additional results of keyframes generation by FusionFrames.}}
    \label{fig:more_results_1}
\end{figure*}

\begin{figure*}[ht]
    \center{\includegraphics[scale=0.44]{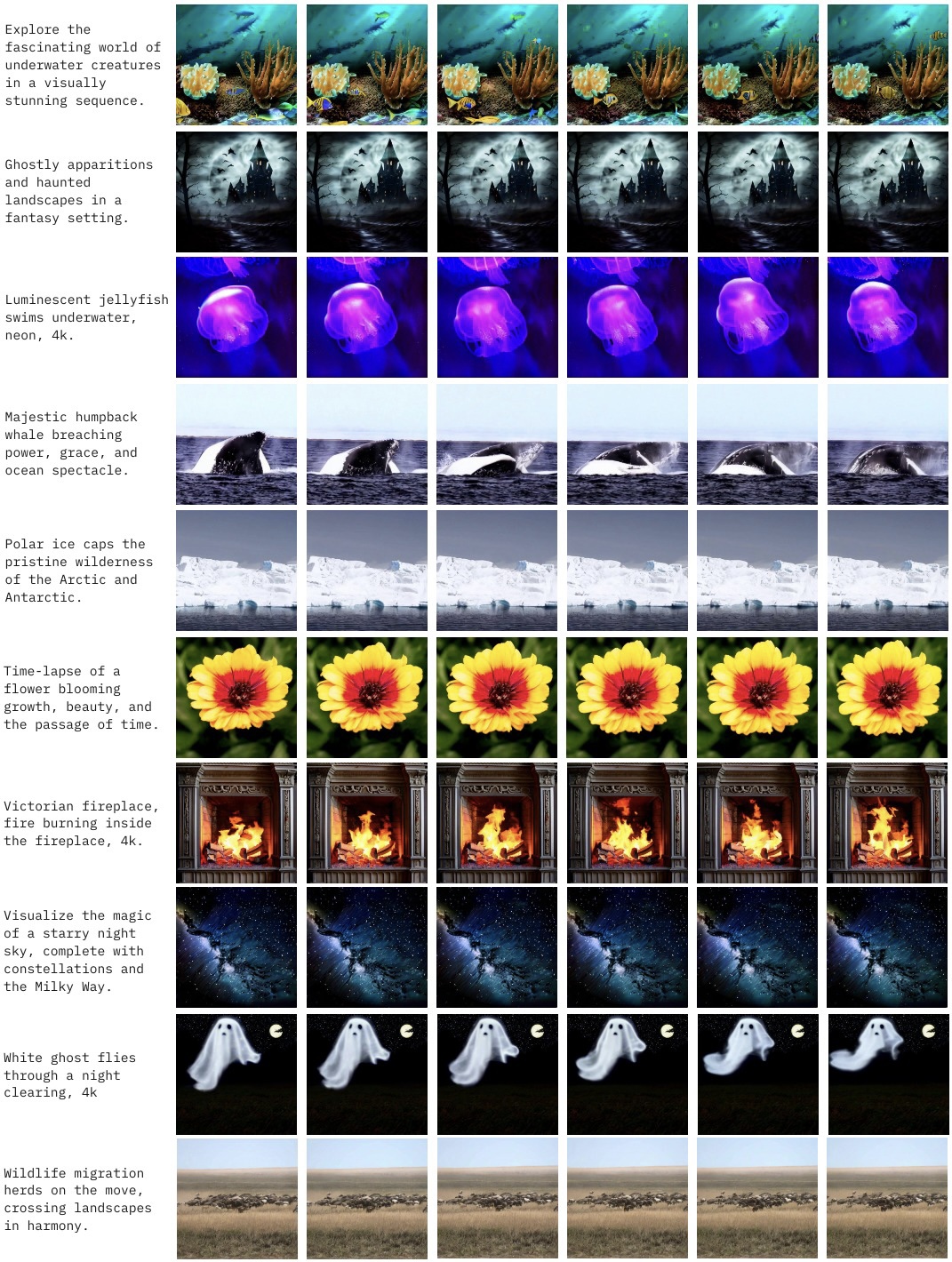}}
    \caption{\textbf{Additional results of keyframes generation by FusionFrames.}}
    \label{fig:more_results_2}
\end{figure*}

\end{document}